\documentclass[lettersize,journal]{IEEEtran}
\usepackage{amsmath,amsfonts}
\usepackage{array}
\usepackage[caption=false,font=normalsize,labelfont=sf,textfont=sf]{subfig}
\usepackage{textcomp}
\usepackage{stfloats}
\usepackage{url}
\usepackage{verbatim}
\usepackage{graphicx}
\usepackage{hyperref}
\usepackage{algorithm}
\usepackage{algpseudocode}
\usepackage{listings}
\usepackage{xcolor}
\usepackage{tikz}
\usepackage{tikz-cd}
\usepackage{hyperref}

\hypersetup{
    colorlinks=true,    % Enable colored links
    linkcolor=blue,     % Color of internal links (e.g., sections, equations)
    citecolor=blue,     % Color of citation links
    urlcolor=blue       % Color of external URLs
}

\lstdefinestyle{mystyle}{
    backgroundcolor=\color{blue!3},   
    commentstyle=\color{green!50!black},
    keywordstyle=\color{blue},
    numberstyle=\tiny\color{gray},
    stringstyle=\color{red!50!black},
    basicstyle=\ttfamily\footnotesize,
    breakatwhitespace=false,         
    breaklines=true,                 
    captionpos=b,                    
    keepspaces=true,                 
    numbers=left,                    
    numbersep=3pt,                  
    showspaces=false,                
    showstringspaces=false,
    showtabs=true,                  
    tabsize=1
}
\newtheorem{definition}{Definition}
\newtheorem{remark}{Remark}

\def\BibTeX{{\rm B\kern-.05em{\sc i\kern-.025em b}\kern-.08em
    T\kern-.1667em\lower.7ex\hbox{E}\kern-.125emX}}
\usepackage{balance}
\begin{document}

\title{Neural Manifolds and Cognitive Consistency:\\ A New Approach to Memory Consolidation in Artificial Systems}

\author{Phuong-Nam Nguyen\\
Faculty of Information Technology\\
School of Technology\\
National Economics University\\
Hanoi, 100000, Vietnam

\thanks{This research is supported by G.A.I.A QTech LLC, Vietnam}

}

\markboth{Journal of \LaTeX\ Class Files,~Vol.~18, No.~9, September~2020}%
{How to Use the IEEEtran \LaTeX \ Templates}

\maketitle

\begin{abstract}
    We introduce a novel mathematical framework that unifies neural population dynamics, hippocampal sharp wave-ripple (SpWR) generation, and cognitive consistency constraints inspired by Heider’s theory. Our model leverages low-dimensional manifold representations to capture structured neural drift and incorporates a balance energy function to enforce coherent synaptic interactions, effectively simulating the memory consolidation processes observed in biological systems. Simulation results demonstrate that our approach not only reproduces key features of SpWR events but also enhances network interpretability. This work paves the way for scalable neuromorphic architectures that bridge neuroscience and artificial intelligence, offering more robust and adaptive learning mechanisms for future intelligent systems. The code repository for this work is available at: \url{git@github.com:namnguyen0510/Neural-Manifolds-and-Cognitive-Consistency.git}.
\end{abstract}


\tableofcontents

\begin{figure*}[h]
    \centering
    \includegraphics[width=\linewidth]{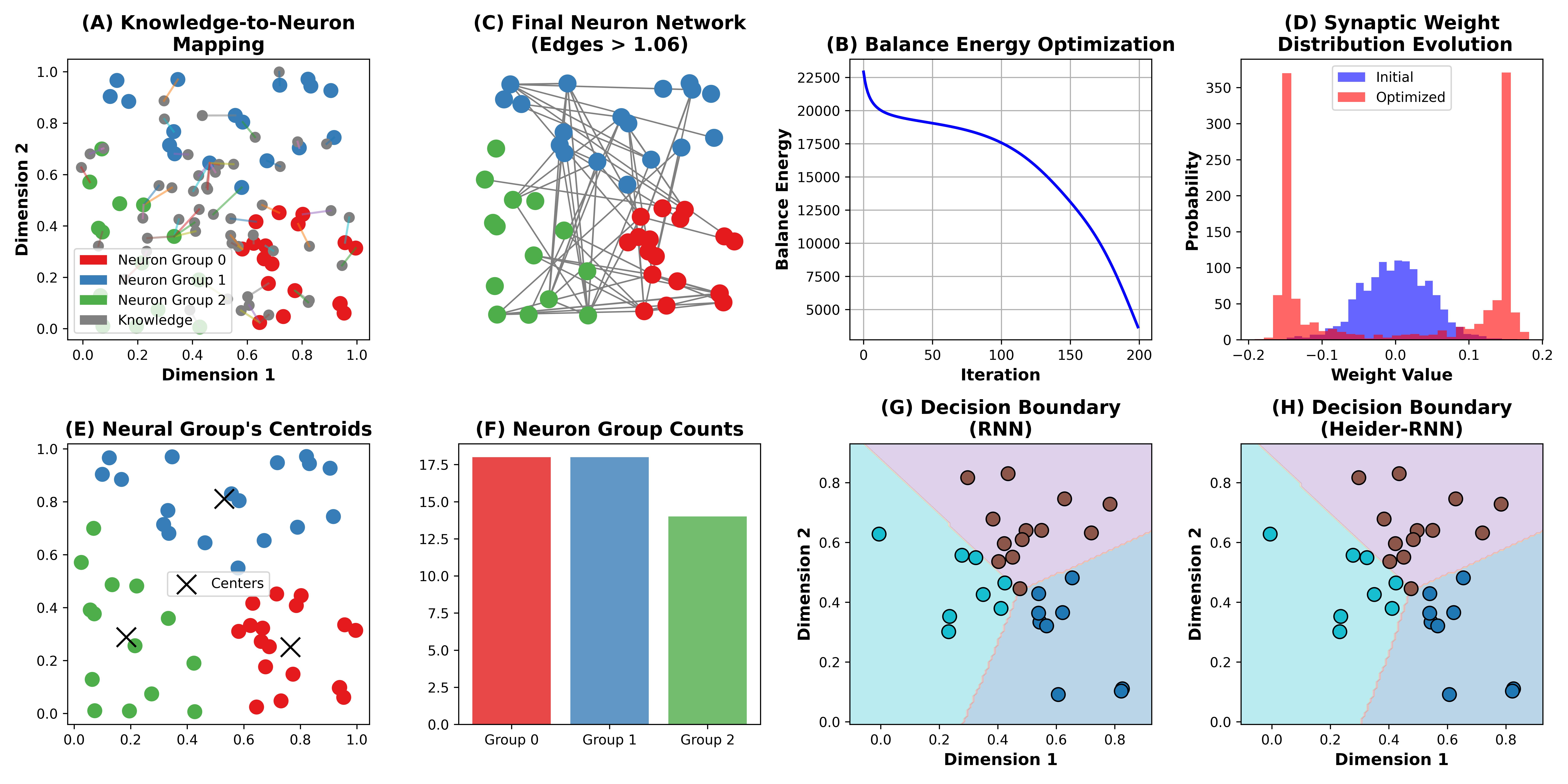}
    \caption{\textbf{Interpretation of neural networks rnn}}
    \label{fig:result_2_2}
\end{figure*}
\section{Introduction} \label{sec:intro}
Understanding and emulating the extraordinary efficiency of the human brain is a grand challenge that continues to inspire advances across neuroscience and artificial intelligence. In recent years, neuromorphic computing\cite{indiveri2011neuromorphic,merolla2014million,izhikevich2006polychronization,davies2018loihi} has emerged as a promising paradigm by designing hardware and algorithms that mimic biological neural dynamics, potentially transforming how machines learn, adapt, and store information. Among the diverse neurobiological phenomena, hippocampal sharp wave-ripple (SpWR) events have garnered significant attention\cite{buzsaki2015hippocampal,yang2024selection}. These brief, high-frequency oscillatory bursts play a crucial role in memory consolidation by facilitating the offline replay of experiences—a process that has inspired innovative strategies for stabilizing learning and mitigating catastrophic forgetting in artificial systems.

Despite extensive research on SpWR and its implications for memory, several critical gaps remain. Current models often isolate the mechanistic aspects of memory replay without adequately incorporating the cognitive consistency that underpins robust neural interactions. In particular, the integration of principles such as Heider’s consistency theory into a unified computational framework is underexplored. Moreover, many existing approaches lack efficient, scalable methods to simulate these complex interactions within large-scale neural architectures. To address these challenges, we propose a novel mathematical framework that unifies neural population dynamics, SpWR generation, and cognitive consistency constraints. Our method leverages a low-dimensional manifold representation to capture the structured drift of neural activity, while a balance energy function enforces coherent inter-neuronal interactions. This integration not only provides a deeper understanding of how memory consolidation emerges from discrete neural events but also offers a scalable computational approach applicable to both biological and artificial neural networks.

Our contributions are threefold. First, we introduce a comprehensive formalism that bridges neurobiological mechanisms and cognitive theories, offering new insights into the interplay between memory replay and neural consistency. Second, we develop efficient algorithms for simulating the proposed framework, demonstrating its potential for scalable neuromorphic implementations. Finally, our empirical evaluations reveal that incorporating cognitive consistency principles enhances both the interpretability and robustness of neural network models, paving the way for more reliable and adaptive artificial intelligence systems.

\section{Backgrounds and Related Works} \label{sec:backgrounds}
\subsection{Sharp Wave-Ripple (SpWR) and neuromorphic computing} \label{subsec:spwr}
Sharp wave‐ripples (SpWR) are brief, high‐frequency oscillatory events observed predominantly in the hippocampus during quiet wakefulness and slow‐wave sleep. First characterized in detail in rodent studies, these transient bursts of synchronized neuronal activity have been implicated in memory consolidation and planning\cite{buzsaki2015hippocampal,yang2024selection}. The SpWR phenomenon has since inspired a host of interdisciplinary investigations, ranging from fundamental neurophysiological studies to the design of brain‐inspired computational systems and machine learning algorithms.

The remarkable efficiency and robustness of SWR-mediated memory consolidation have motivated researchers to incorporate similar mechanisms into neuromorphic systems—hardware architectures designed to emulate biological neural networks. Neuromorphic computing leverages spiking neural networks (SNNs), where information is encoded in discrete spikes rather than continuous activations, mirroring the event-driven nature of biological neurons. \cite{indiveri2011neuromorphic,merolla2014million} have pioneered neuromorphic circuits that mimic key dynamical properties of biological networks. In these systems, incorporating transient high-frequency events akin to SWRs can promote offline “replay” of spatiotemporal patterns, thereby enabling synaptic modifications that resemble biological learning. For instance, \cite{izhikevich2006polychronization} works on polychronization demonstrated that networks of spiking neurons could produce complex time-locked patterns, offering a computational parallel to the sequence reactivation observed during hippocampal ripples. Furthermore, neuromorphic platforms such as Intel’s Loihi processor\cite{davies2018loihi} support on-chip learning rules that can benefit from replay-like dynamics. By emulating SWR-inspired memory consolidation, these systems can potentially overcome challenges like catastrophic forgetting—a common limitation in conventional artificial neural networks.

The translation of SWR mechanisms into the domain of machine intelligence has been most notably influential in reinforcement learning (RL) paradigms. The concept of “experience replay,” as popularized in deep RL algorithms\cite{mnih2015human}, is loosely inspired by the hippocampal replay observed during SWRs. In these algorithms, past experiences are sampled and reprocessed to stabilize learning and improve sample efficiency—an idea conceptually analogous to the memory consolidation functions of SWRs in the brain. Moreover, theoretical frameworks like complementary learning systems suggest that integrating fast, hippocampus-like memory systems with slower, neocortex-inspired networks can enhance the ability of artificial agents to generalize from limited data\cite{kumaran2016learning}. Recent advances in spiking deep networks further indicate that incorporating temporal dynamics, including SWR-like events, may lead to more robust and energy-efficient machine intelligence\cite{kim2019response}. The cross-pollination between neuroscience and machine intelligence not only enriches our theoretical understanding but also inspires practical algorithms that leverage the brain’s natural strategies for dealing with sequential data, prediction, and memory consolidation.

\subsection{Heider's consistent theory} \label{subsec:heider}
Heider’s consistency theory\cite{heider2013psychology}, commonly known as Balance Theory, posits that individuals are inherently motivated to maintain coherence among their cognitions, emotions, and interpersonal relationships. In situations where inconsistencies arise, individuals experience psychological discomfort and are driven to restore equilibrium.

\begin{definition}[The Drive for Consistency] Individuals demonstrate a preference for consistency across their attitudes, beliefs, and relationships. When discrepancies occur, the resulting discomfort compels them to seek resolution. \end{definition}

The foundational model of the theory employs a tripartite structure comprising the following elements: \begin{enumerate} \item \textbf{P}: a person (e.g., oneself), \item \textbf{O}: another person (e.g., a friend), \item \textbf{X}: an object, idea, or attitude (e.g., a brand or belief). \end{enumerate}

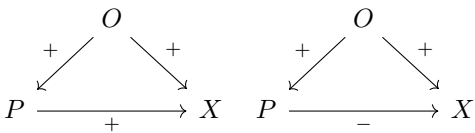
\begin{figure}[h] 
\centering 
\begin{tikzcd}[row sep=2em, column sep=2em]
& O \arrow[dl, "+", swap] \arrow[dr, "+"] & \\
P \arrow[rr, "+", swap] & & X
\end{tikzcd}
\begin{tikzcd}[row sep=2em, column sep=2em]
& O \arrow[dl, "+", swap] \arrow[dr, "+"] & \\
P \arrow[rr, "-", swap] & & X
\end{tikzcd}
\caption{Graphical representation of Heider’s consistency theory: (Left) represents a balanced triad; (Right) represents an imbalanced triad.} 
\end{figure}

In this model, each dyadic relationship (i.e., P–O, O–X, and P–X) may be positive (indicating liking or approval) or negative (indicating disliking or disapproval). A triadic configuration is deemed balanced when the overall relational sentiments are harmonious. For example, if an individual (P) likes a friend (O) and that friend holds a favorable view of a particular movie (X), the individual is more inclined to also view the movie positively to maintain balance. Conversely, an imbalance arises when inconsistency is present; for instance, if an individual (P) likes a friend (O) but dislikes a movie (X) that the friend favors, this discrepancy generates psychological tension. To alleviate this tension, the individual may modify their opinion of the movie or attempt to influence the friend’s perspective, thereby reestablishing equilibrium.

\section{The Proposed Framework} \label{sec:framework}
\subsection{Mathematical formalism of SpWR} \label{subsec:math_spwr}
To model SpWR events and their role in memory consolidation, we propose a mathematical framework that integrates \textit{neural population dynamics}, \textit{sequence replay}, and \textit{trial block selection}. 

\begin{definition}[\textbf{Neural population dynamics}] 
    The collective activity of a group of neurons, typically described in terms of firing rates, spike correlations, or low-dimensional representations of neural states. In the context of the proposed model, neural population dynamics refer to how the firing patterns of place cells evolve over time, influenced by external inputs (e.g., an agent's position in a maze\cite{yang2024selection}) and internal factors (e.g., synaptic plasticity and structured drift).
\end{definition}
\begin{definition}[\textbf{Sequence replay}]
    The reactivation of temporally structured neural activity patterns that resemble those observed during wakeful experiences. Sequence replay often occurs during quiescent states, particularly in SpWR events, and is thought to support memory consolidation\footnote{Memory consolidation is the process of turning short-term memories into long-term memories. It's a time-dependent process that involves structural and chemical changes in the brain.} by reinforcing neural representations of past experiences.
\end{definition}
\begin{definition}[\textbf{Trial Block Selection}]
    The process by which specific subsets of past experiences (trial blocks) are preferentially reactivated during replay. In this model, trial block selection is governed by a probabilistic rule that favors reactivation of experiences that were strongly activated during wakefulness, modulated by similarity metrics and weighting parameters that influence memory consolidation.
\end{definition}

\subsubsection{Neural population dynamics}\label{subsubsec:neural_dynamics}
During exploration (theta state), place cells\footnote{A \textit{place cell} is a type of neuron found in the hippocampus that becomes active when an animal is in a specific location within its environment. Each place cell has a preferred spatial location, known as its place field, where it fires action potentials at a high rate. Place cells collectively form a cognitive map, allowing animals to navigate and encode spatial memories.} exhibit drifting receptive fields\footnote{A \textit{receptive field} refers to the specific region of sensory space (e.g., visual field, auditory frequency, or spatial position) where a neuron responds to stimuli. In the context of place cells, the receptive field is called the place field, which is the area in the environment where the neuron exhibits elevated firing activity. The location of a place field can shift over time due to factors such as experience, learning, and neural plasticity.}. Let $\mathbf{r}_t \in \mathbb{R}^N$ represent the firing rates of $N$ neurons at time $t$. The population activity evolves as
\begin{equation}\label{eq:neural_activity}
    \mathbf{r}_t = \mathbf{f}(\mathbf{x}_t) + \mathbf{\epsilon}_t,
\end{equation}
where $\mathbf{x}_t$ is the agent's position in the maze; $\mathbf{f}(\cdot)$ maps position to firing rates (place fields); and $\mathbf{\epsilon}_t \sim \mathcal{N}(0, \sigma^2)$ models trial-to-trial variability (drift).

The drift is structured (non-random), as shown by embeddings on manifolds and decoding accuracy. This is captured by a low-dimensional manifold $\mathcal{M} \subset \mathbb{R}^N$, parameterized by position $\mathbf{x}$ and trial block $k$
\begin{equation}\label{eq:manifold}
    \mathcal{M} = \left\{ \mathbf{r}(\mathbf{x}, k) \mid \mathbf{x} \in \text{maze}, \, k \in \{1, 2, \dots, K\} \right\}.
\end{equation}
Here, we use the term maze to appreciate the original work\cite{yang2024selection}; however, this term can be generalized to any environment in which the agent resides.
\begin{remark}
    In the context of Agentic AI and neural networks, the structured (non-random) drift refers to how an agent's internal representations evolve in a predictable manner rather than through arbitrary fluctuations. This drift is evident in manifold embeddings, where high-dimensional neural activity patterns can be projected onto a low-dimensional space that retains meaningful structure (good representation\cite{bengio2013representation}). While the term maze originates from neuroscience studies involving spatial navigation tasks, this concept extends to any environment in which the agent operates—whether it is a robotic system navigating real-world terrain, a reinforcement learning agent solving a game, or a language model adapting to conversational context. The manifold structure captures how an agentic AI organizes its internal state representations over time, facilitating efficient decision-making and memory consolidation.
\end{remark}

\subsubsection{SpWR Generation}\label{subsubsec:spwr_generation}
SpWR occur during immobility (e.g., reward consumption)\footnote{SpWR events can be interpreted as \textit{offline memory reactivation} that occurs when an agent is temporarily \textit{inactive}, such as during reward consumption or decision pauses.}. Let $\mathbf{s}_m \in \mathbb{R}^N$ denote the spike content of the $m$-th SpWR. The probability of replaying trial block $k$ depends on its activation during wakefulness:
\begin{equation}\label{eq:spwr_naive}
    P(k \mid m) \propto \exp\left( \beta \cdot \text{sim}(\mathbf{s}_m, \mathbf{r}_k) \right),
\end{equation}
where $\text{sim}(\cdot, \cdot)$ measures similarity (e.g., cosine similarity) between SpWR activity $\mathbf{s}_m$ and trial block $k$'s template $\mathbf{r}_k$ and $\beta$ controls the sharpness of selection (higher $\beta$ = stronger bias toward salient trials).

\subsubsection{Replay content and memory consolidation}\label{subsubsec:replay_activation}
SpWR replay sequences from the manifold $\mathcal{M}$. Let $\mathbf{z}_m \in \mathcal{M}$ be the low-dimensional embedding of $\mathbf{s}_m$. The replay trajectory $\mathbf{z}_m(t)$ during SpWR $m$ is modeled as a diffusion process on $\mathcal{M}$
\begin{equation}
    d\mathbf{z}_m(t) = \mathbf{A} \cdot \mathbf{z}_m(t) \, dt + \mathbf{B} \, d\mathbf{W}(t),
\end{equation}
where $\mathbf{A}$ and $\mathbf{B}$ are matrices governing drift and diffusion and $\mathbf{W}(t)$ is a Wiener process. Decoding trial blocks\footnote{\textit{Decoding trial blocks} refers to the process of determining which past experience or experimental condition an observed neural activity pattern corresponds to.} from $\mathbf{z}_m(t)$ uses a unsupervised classifier $\mathcal{D}$ (e.g., k-nearest neighbors):
\begin{equation}\label{eq:neural_clustering}
    \hat{k}_m = \mathcal{D}(\mathbf{z}_m(t)).
\end{equation}

The \textit{wake-sleep interaction} describes how neural activity during wakefulness influences memory consolidation during sleep. This process involves SpWR events, which replay past experiences to strengthen important memories (memory consolidation).  

During \textit{wakefulness}, SpWR events occur, selectively activating trial blocks (distinct past experiences). The probability of a specific trial block $k$ being replayed during wakefulness is denoted as $P(k \mid \text{wake})$.  This probability biases subsequent sleep SpWRs, meaning that experiences frequently replayed during wakefulness are more likely to be replayed during sleep. Mathematically, this is expressed as:
\begin{equation}\label{eq:wakefulness}
    P(k \mid \text{sleep}) \propto P(k \mid \text{wake})^\gamma,
\end{equation}
where $\gamma > 1$ amplifies the selection effect. \textit{This means that strongly activated memories during wakefulness become even more likely to be replayed during sleep, reinforcing their consolidation}.

Memory consolidation also involves \textit{synaptic plasticity}, where neural connections are strengthened based on replay activity. This process is modeled using a weight update rule:
\begin{equation}\label{eq:synaptic_plasticity}
    \Delta w_{ij} \propto \eta \cdot \sum_{m} P(k \mid m) \cdot \mathbf{s}_m^{(i)} \mathbf{s}_m^{(j)}
\end{equation}
where $\eta$ is the learning rate, controlling how much synapses are updated; $\mathbf{s}_m^{(i)}$ is the activity of neuron $i$ during SpWR $m$; $P(k \mid m)$ is the probability of trial block $k$ given the SpWR $m$; and $\mathbf{s}_m^{(i)} \mathbf{s}_m^{(j)}$ represents synaptic co-activation, strengthening neural connections. \textit{This means that when a trial block is replayed during SpWRs, the synapses associated with that experience strengthen, making future retrieval easier}.

\begin{remark}
    This weight update rule is analogous to Hebbian learning ("neurons that fire together, wire together") and is used in AI models for associative memory and reinforcement learning. Their are three key-takes from our proposed-mathematical formulation: (1) Experiences replayed more often during wakefulness are more likely to be replayed during sleep; (2) Sleep replay strengthens synaptic connections through synaptic tagging; and (3) This process helps consolidate important memories while forgetting less relevant ones.  
\end{remark}

The remaining questions are: (1) how neural representations evolve over time? and (2) how SpWR events are probabilistically associated with past experiences? For the former question, we adopt the \textit{manifold dynamics}, given as
\begin{equation}\label{eq:manifold_dyna}
    \mathbf{z}_{k+1} = \mathbf{z}_k + \mathbf{v} \Delta t + \mathbf{\xi}_k
\end{equation}
where $\mathbf{z}_k$ represents the low-dimensional neural state at trial block $k$, embedded on a manifold $\mathcal{M}$ (a structured representation of neural activity); $\mathbf{v}$ is the drift velocity, representing gradual changes in neural activity over time; and $\mathbf{\xi}_k$ is random noise, modeled as $\mathcal{N}(0, \Sigma)$\footnote{A Gaussian distribution with covariance $\Sigma$.}.  
\begin{remark}
    Equation~\ref{eq:manifold_dyna} models how neural representations shift over time, capturing structured drift rather than purely random fluctuations. The drift term ($\mathbf{v} \Delta t$) accounts for slow representational changes, possibly due to learning or experience accumulation. The noise term ($\mathbf{\xi}_k$) represents trial-to-trial variability, adding randomness to the process.
\end{remark}

For the latter question, we derive the SpWR likelihood function
\begin{equation}\label{eq:sprw_likelihood}
    \log P(\mathbf{s}_m \mid k) = -\frac{1}{2} \left\| \mathbf{s}_m - \mathbf{r}_k \right\|_{\Sigma^{-1}}^2 + C
\end{equation}
where $P(\mathbf{s}_m \mid k)$ is the probability that SpWR $m$ corresponds to trial block $k$; $\mathbf{s}_m$ is the neural activity during the SpWR event; $\mathbf{r}_k$ is the template activity of trial block $k$ (i.e., the stored memory representation); $\|\mathbf{s}_m - \mathbf{r}_k\|_{\Sigma^{-1}}^2$ is a Mahalanobis distance, measuring the similarity between SpWR activity and trial block templates, weighted by $\Sigma^{-1}$ (the inverse covariance matrix); $C$ is a normalization constant ensuring the probability distribution sums to 1.
\begin{remark}
    Equation~\ref{eq:sprw_likelihood} defines a statistical model where SpWR events are probabilistically linked to past experiences. The likelihood is higher when $\mathbf{s}_m$ closely resembles $\mathbf{r}_k$, meaning SpWR are more likely to replay memories with similar neural activity. The covariance $\Sigma$ accounts for variability in neural activity, allowing a probabilistic match rather than a strict one. 
\end{remark}

\subsection{Principles of cognitive consistency in neural interactions} \label{subsec:cognitive_consistency}
To integrate Heider's consistency theory into the mathematical model of SpWR dynamics, we extend the neural population framework to incorporate principles of \textit{cognitive consistency} (Heider's consistency theory, Section~\ref{subsec:heider}) in neural interactions. This bridges social network balance with synaptic and ensemble relationships in the hippocampus.
\subsubsection{Neural population dynamics with balance constraints}
We define a \textit{signed graph} $\mathcal{G} = (\mathcal{N}, \mathcal{E})$, where:
\begin{itemize}
    \item Nodes $\mathcal{N}$ represent neurons or neural ensembles.
    \item Edges $\mathcal{E}$ encode pairwise interactions (positive/negative) via synaptic weights $W_{ij} \in \mathbb{R}$.
\end{itemize}
We construct a balance energy function $E_{\text{balance}}$ to penalize imbalanced triads (triplets of neurons):
\begin{equation}\label{eq:e_balance}
    E_{\text{balance}} = \sum_{i<j<k} \left( W_{ij}W_{jk}W_{ki} + 1 \right)^2,
\end{equation}
where the sum is over all triads. Balanced triads ($W_{ij}W_{jk}W_{ki} > 0$) have low energy; imbalanced triads ($W_{ij}W_{jk}W_{ki} < 0$) have high energy. We modify the population activity in Equation~\ref{eq:neural_activity} to include balance-driven stabilization
\begin{equation}
    \mathbf{r}_t = \mathbf{f}(\mathbf{x}_t) - \alpha \nabla_{\mathbf{r}} E_{\text{balance}} + \mathbf{\epsilon}_t,
\end{equation}
where $\alpha$ controls the strength of balance enforcement. This drives neural activity toward configurations with balanced interactions.

\subsubsection{SpWR generation with balance-weighted replay}
We adjust the SpWR replay probability (Equation~\ref{eq:spwr_naive}) to favor trial blocks with balanced neural interactions
\begin{equation}
    P(k \mid m) \propto \exp\left( \beta \cdot \text{sim}(\mathbf{s}_m, \mathbf{r}_k) - \gamma E_{\text{balance}}^{(k)} \right),
\end{equation}
where $E_{\text{balance}}^{(k)}$ is the balance energy of trial block $k$, and $\gamma$ weights the importance of balance. We update synaptic weights to prioritize balanced interactions during replay
\begin{equation}
    \Delta W_{ij} \propto \eta \sum_{m} P(k \mid m) \cdot \left( \mathbf{s}_m^{(i)} \mathbf{s}_m^{(j)} - \lambda \frac{\partial E_{\text{balance}}}{\partial W_{ij}} \right),
\end{equation}
where $\lambda$ governs the trade-off between Hebbian plasticity and balance enforcement. Finally, we modify the formularized Equation~\ref{eq:manifold_dyna} and ~\ref{eq:sprw_likelihood} as \textit{balance-aware} manifold dynamics:
\begin{equation}\label{eq:balanced_dynamics}
\boxed{
    \begin{split}
        \mathbf{z}_{k+1} &= \mathbf{z}_k + \mathbf{v}\Delta t - \alpha \nabla_{\mathbf{z}} E_{\text{balance}} + \mathbf{\xi}_k\\
        \log P(\mathbf{s}_m \mid k) &= -\frac{1}{2} \left\| \mathbf{s}_m - \mathbf{r}_k \right\|_{\Sigma^{-1}}^2 - \gamma E_{\text{balance}}^{(k)} + C
    \end{split}
    }.
\end{equation}
\begin{remark}
     The manifold dynamics in Equation~\ref{eq:balanced_dynamics} reflect a balance-driven “\textit{structured drift}” where the system is nudged toward regions of high internal consistency (i.e., low \(E_{\text{balance}}\)), facilitating stable memory representations. Besides, the proposed model also predicts that neural states with lower balance energy\footnote{Smaller energy means more balanced or consistent} and greater sensory congruence are preferentially consolidated.
\end{remark}
\begin{remark}
    The proposed framework suggests that neural balance constraints shape memory dynamics by enforcing structured, stable representations. Attractor dynamics emerge as low-energy configurations, ensuring cognitive consistency and resistance to perturbations. Selective replay mechanisms during SpWR events prioritize experiences with balanced interactions, leading to preferential consolidation of internally consistent memories. Decoding accuracy improves as neural manifolds minimize imbalance, supporting more reliable sensory representation. Finally, synaptic plasticity signatures indicate that balanced triads exhibit stronger LTP, offering a potential biomarker for effective memory encoding. These insights provide testable predictions for future neurophysiological and computational studies.
\end{remark}

\subsection{Multi-expert modeling with cognitive consistency} \label{subsec:multi_expert}
\subsubsection{Agent knowledge and neural mapping}
Let an agent possess a set of knowledge items:
\begin{equation}
    \vec{K} = \bigl( K_1, K_2, \dots, K_n \bigr)
\end{equation}
and let \(\vec{N} = \bigl( N_1, N_2, \dots, N_m \bigr)\) denote a set of neurons (or neural ensembles) that are available to represent this knowledge. We assume a mapping:
\begin{equation}
    \phi: \vec{K} \to \vec{N},
\end{equation}
which assigns each knowledge item \(K_i\) to one or more neurons. In many settings, this mapping is defined through an embedding. For instance, if there is an embedding function:
\begin{equation}
    \psi: \vec{K} \to \mathbb{R}^d,
\end{equation}
and each neuron \(N_j\) is associated with a feature vector \(v_j \in \mathbb{R}^d\), one natural choice is:
\begin{equation}\label{eq:neural_maps}
    \phi(K_i) = N_{j^*} \quad \text{with} \quad j^* = \arg\min_{j=1,\dots,m} \| \psi(K_i) - v_j \|.
\end{equation}

\subsubsection{Synaptic interactions and balance}
Let the pairwise interaction between neurons be encoded in a synaptic weight matrix
\begin{equation}]\label{eq:synatic_weight}
    W = \left[ W_{ij} \right] \in \mathbb{R}^{m \times m},
\end{equation}
where \(W_{ij}\) can be positive (excitatory) or negative (inhibitory). According to Heider’s balance theory, triads (sets of three neurons) should ideally be balanced. For any triad \((N_i, N_j, N_k)\), the product:
\begin{equation}
    W_{ij} \, W_{jk} \, W_{ki}
\end{equation}
indicates the balance of the triad. A triad is considered balanced if
\begin{equation}
    W_{ij} \, W_{jk} \, W_{ki} > 0,
\end{equation}
and imbalanced otherwise (discussed in the previous section).
\subsubsection{Synaptic plasticity and network dynamics}
The agent’s neural system tends to adjust synaptic strengths to reduce cognitive dissonance. This dynamic can be modeled by gradient descent on the balance energy
\begin{equation}
    \frac{dW_{ij}}{dt} = -\eta \, \frac{\partial E_{\text{balance}}}{\partial W_{ij}},
\end{equation}
where \(\eta > 0\) is a learning rate. This equation drives the network toward configurations that minimize \(E_{\text{balance}}\), thus promoting internal consistency in the agent’s knowledge representation.

\subsubsection{Algorithms}
The algorithm for computing the balance energy is:
\begin{lstlisting}[language=Python, caption=Computation of balance energy]
def compute_balance_energy(W):
    N = W.shape[0]
    energy = 0.0
    for i in range(N):
        for j in range(i+1, N):
            for k in range(j+1, N):
                term = W[i, j] * W[j, k] * W[k, i] + 1
                energy += term2
    return energy
\end{lstlisting}

The gradient of the balance energy with respect to \( W \) is given by:
\begin{equation}\label{eq:e_diff}
\frac{\partial E}{\partial W_{ij}} = 2 \left( W_{ij} W_{jk} W_{ki} + 1 \right) W_{jk} W_{ki}.
\end{equation}
Similarly, the contributions for other matrix entries are:
\begin{align}
\frac{\partial E}{\partial W_{jk}} &= 2 \left( W_{ij} W_{jk} W_{ki} + 1 \right) W_{ij} W_{ki}, \\
\frac{\partial E}{\partial W_{ki}} &= 2 \left( W_{ij} W_{jk} W_{ki} + 1 \right) W_{ij} W_{jk}.
\end{align}
The algorithm for computing the gradient is:
\begin{lstlisting}[language=Python, caption=Computation of balance energy gradient]
def compute_balance_gradient(W):
    N = W.shape[0]
    grad = np.zeros_like(W)
    for i in range(N):
        for j in range(i+1, N):
            for k in range(j+1, N):
                triad_val = W[i, j] * W[j, k] * W[k, i] + 1
                factor = 2 * triad_val
                # Contribution for W[i,j]
                grad[i, j] += factor * (W[j, k] * W[k, i])
                grad[j, i] += factor * (W[j, k] * W[k, i])
                # Contribution for W[j,k]
                grad[j, k] += factor * (W[i, j] * W[k, i])
                grad[k, j] += factor * (W[i, j] * W[k, i])
                # Contribution for W[k,i]
                grad[k, i] += factor * (W[i, j] * W[j, k])
                grad[i, k] += factor * (W[i, j] * W[j, k])
    return grad
\end{lstlisting}

\subsubsection{Improving the computational efficiency}
Clearly, both Algorithm 1 and 2 for balance energy and gradient requires $\mathcal{O}(N^3)$, which is not good for model scalability. We can greatly speed up these computations by expressing the triple‐summations in terms of matrix traces. In particular, for a symmetric weight matrix \(W\), note that $E_{\text{balance}}$ can be expanded as
\begin{equation}
    E_{\text{balance}} = \underbrace{\binom{N}{3}}_{\text{constant}} + 2\sum_{i<j<k} W_{ij}W_{jk}W_{ki} + \sum_{i<j<k} \Bigl(W_{ij}W_{jk}W_{ki}\Bigr)^2.
\end{equation}
Two key observations make a fast (vectorized) computation possible:
\begin{enumerate}
    \item \textit{Triple product term}: For a symmetric \(W\), it is known that\begin{equation}
        \sum_{i<j<k} W_{ij}W_{jk}W_{ki} = \frac{1}{6}\operatorname{tr}\bigl(W^3\bigr).
    \end{equation}
    \item \textit{Squared triple product term}: Notice that
    \begin{equation}
        \Bigl(W_{ij}W_{jk}W_{ki}\Bigr)^2 = W_{ij}^2\, W_{jk}^2\, W_{ki}^2.
    \end{equation}
    If we define the elementwise square \(A = W\circ W\) (i.e. \(A_{ij}=W_{ij}^2\)), then similarly
    \begin{equation}
        \sum_{i<j<k} W_{ij}^2\, W_{jk}^2\, W_{ki}^2 = \frac{1}{6}\operatorname{tr}\bigl(A^3\bigr).
    \end{equation}
\end{enumerate}
Thus, the energy can be rewritten as
\begin{equation}
    E^\text{fast}_{\text{balance}} = \binom{N}{3} + \frac{1}{3}\operatorname{tr}\bigl(W^3\bigr) + \frac{1}{6}\operatorname{tr}\bigl((W\circ W)^3\bigr).
\end{equation}

A similar derivation works for the gradient. When we differentiate $E^\text{fast}_{\text{balance}}$ the contributions are:
\begin{itemize}
    \item For the \(\operatorname{tr}\bigl(W^3\bigr)\) term: \(\frac{d}{dW}\operatorname{tr}\bigl(W^3\bigr) = 3W^2\), so the derivative gives \(W^2\).
    \item For the \(\operatorname{tr}\bigl((W\circ W)^3\bigr)\) term: Using the chain rule with \(A = W\circ W\) (so that \(dA/dW = 2W\) elementwise) and the fact that
  \(\frac{d}{dA}\operatorname{tr}\bigl(A^3\bigr) = 3A^2\),
  we obtain a gradient contribution of\begin{equation}
      \left( \frac{1}{6}\cdot 3 \cdot 2 \right) \cdot \Bigl[(W\circ W)^2 \circ W\Bigr] = (W\circ W)^2 \circ W.
  \end{equation}
\end{itemize}
Since our derivation using traces counts each triad in a way that gives half the contribution compared to summing explicitly over \(i<j<k\), we multiply the combined result by 2. Thus, the fast gradient is given by:
\begin{equation}
    \text{grad\_fast} = 2\Bigl(W^2 + \Bigl[(W\circ W)^2 \circ W\Bigr]\Bigr).
\end{equation}
Here, \(W^2\) is the usual matrix product \(W @ W\), and \((W\circ W)^2\) is computed as the matrix product of the elementwise squared matrix with itself.

The implementation for the proposed fast computations is given as
\begin{lstlisting}[language=Python, caption=Fast computation of balance energy and gradient]
def compute_balance_energy_fast(W):
    N = W.shape[0]
    comb = N * (N - 1) * (N - 2) / 6  # Number of triads
    energy = comb + np.trace(W @ W @ W) / 3 + np.trace((np.square(W) @ np.square(W) @ np.square(W))) / 6
    return energy

def compute_balance_gradient_fast(W):
    term1 = W @ W
    term2 = np.dot(np.square(W), np.square(W)) * W  # elementwise multiplication after matrix product
    grad_fast = 2 * (term1 + term2)
    return grad_fast
\end{lstlisting}

\section{Results}\label{sec:results}
The code for this evaluation is available at: \url{git@github.com:namnguyen0510/Neural-Manifolds-and-Cognitive-Consistency.git}.
\begin{figure*}[h]
    \centering
    \includegraphics[width=\linewidth]{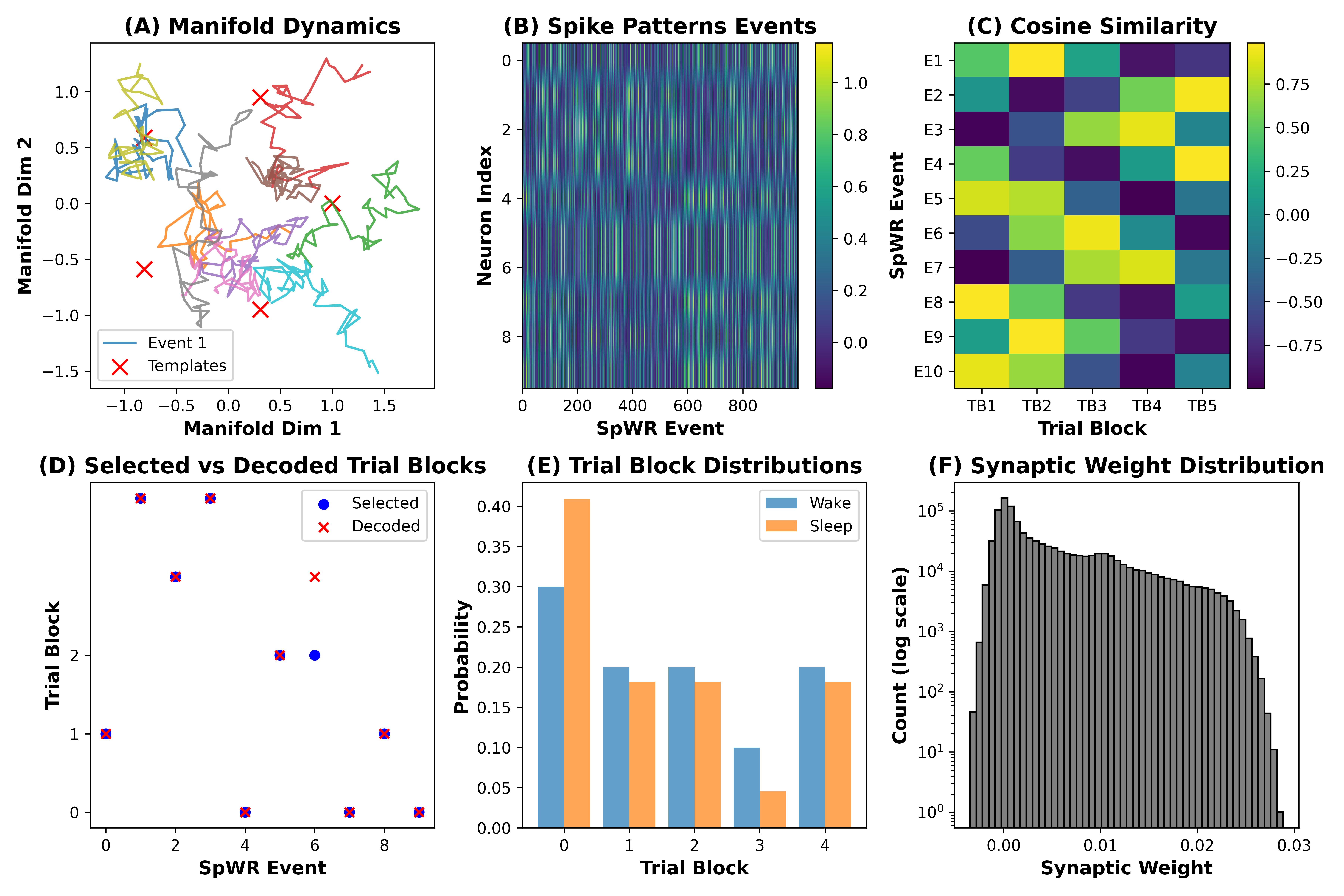}
    \caption{\textbf{Simulation of neural population dynamics and SpWR generation}}
    \label{fig:result_1}
\end{figure*}
\begin{figure*}[h]
    \centering
    \includegraphics[width=\linewidth]{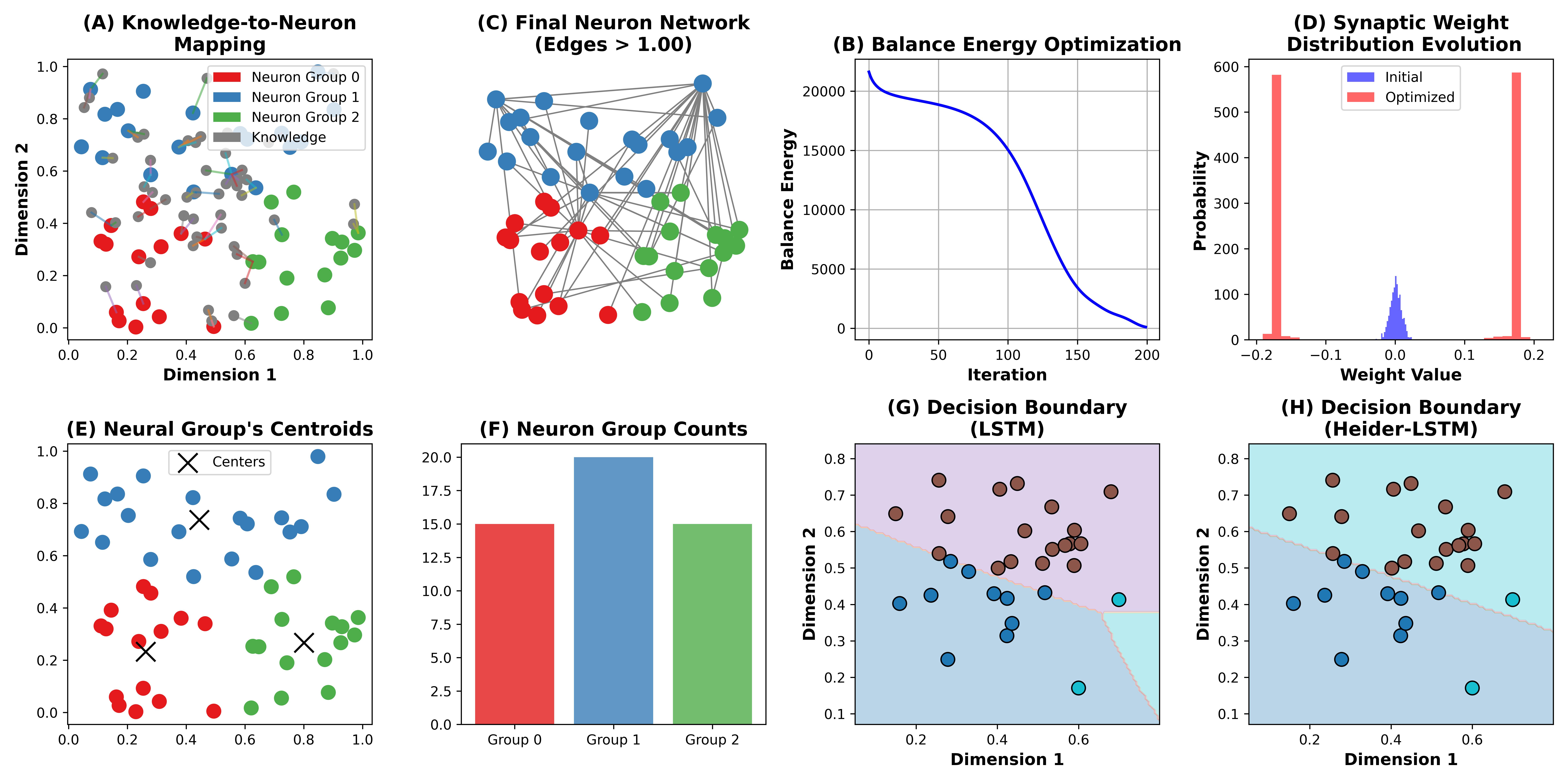}
    \includegraphics[width=\linewidth]{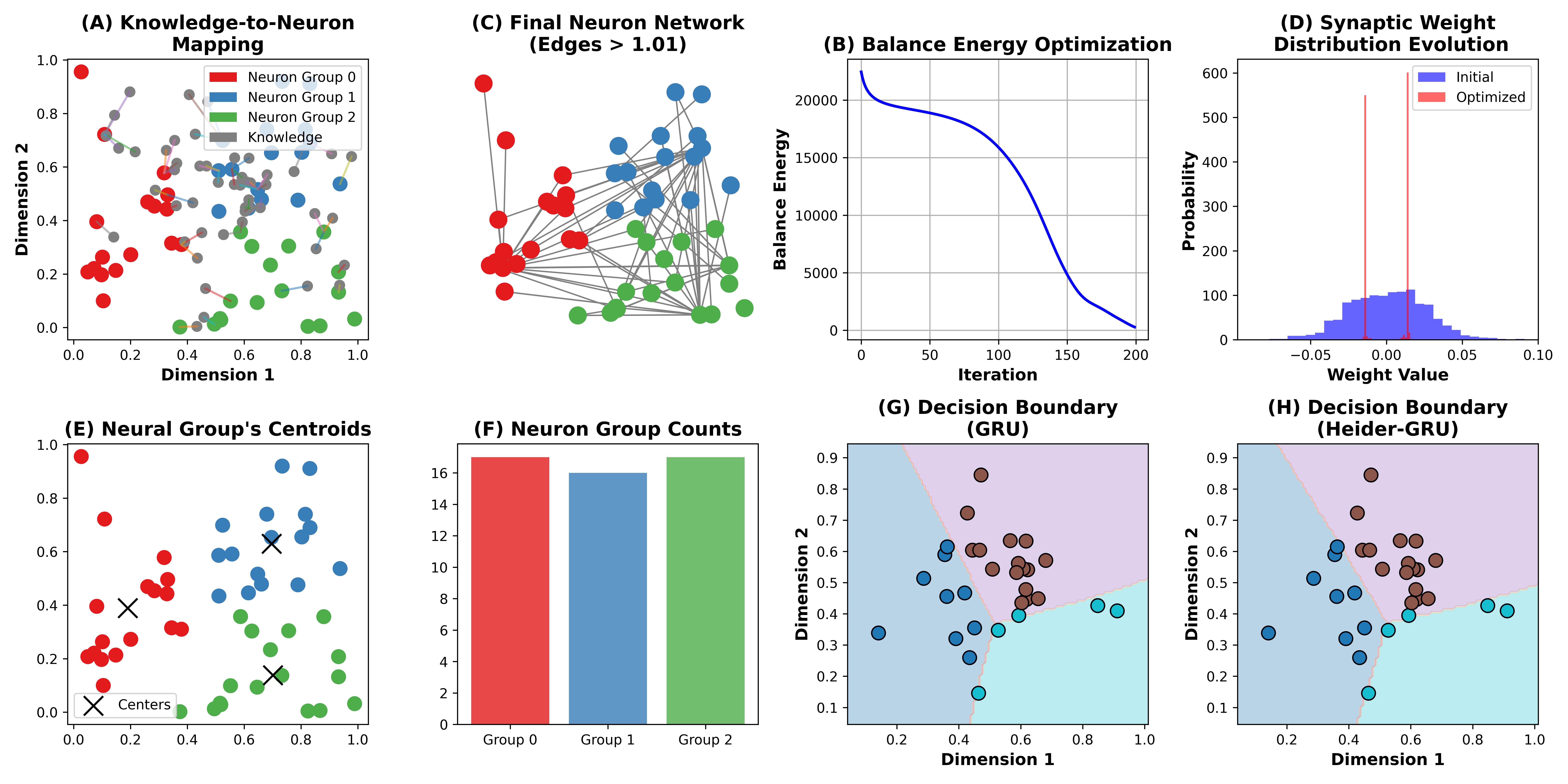}
    \caption{\textbf{Interpretation of neural networks: (Top) LSTM. (Bottom) GRU}}
    \label{fig:result_2_1}
\end{figure*}

\subsection{Simulation of neural population dynamics and SpWR generation}
\noindent \textbf{Configuration space:} A population of \texttt{N = 1000} neurons is initialized, where each neuron has a preferred position sampled uniformly from \([0,1]\). The preferred positions are used to construct place fields, modeled as Gaussian functions:
\begin{equation}
    \text{place\_field}(x, \text{pref}, \sigma) = \exp\left(-\frac{(x - \text{pref})^2}{2\sigma^2}\right)
\end{equation}
where \(\sigma_{\text{pf}}\) is the standard deviation controlling the width of the place field. Trial positions are selected linearly within the range \([0.2, 0.8]\), and corresponding neural activity templates (\texttt{r\_templates}) are computed. To represent trial positions in a low-dimensional space, each trial position is mapped to a 2D manifold using a circular embedding
\begin{equation}
    z_k = \left(\cos\left(\frac{2\pi k}{K}\right), \sin\left(\frac{2\pi k}{K}\right) \right)
\end{equation}
where \(K\) is the number of trial blocks. SpWR events are generated by simulating diffusion in the 2D manifold space. The diffusion process follows
\begin{equation}
    z_{t+1} = z_t + v \cdot dt + \eta
\end{equation}
where \(\eta\) is Gaussian noise with variance proportional to the diffusion coefficient. The similarity between the final state of the trajectory and each trial embedding is computed using cosine similarity:
\begin{equation}
    \text{cosine\_similarity}(a, b) = \frac{a \cdot b}{\|a\| \|b\| + 10^{-10}}
\end{equation}
We add a small term $10^{-10}$ for numerical stability. A softmax function with inverse temperature \(\beta\) is applied to obtain probabilities for selecting trial blocks. Neural activity for SpWR events is sampled from the selected trial block with added Gaussian noise. Hebbian learning is applied to update synaptic weights using:
\begin{equation}
    W \leftarrow W + \eta \cdot p_k \cdot (s_m \otimes s_m)
\end{equation}
where \(s_m\) is the sampled neural activity during SpWR, and \(p_k\) is the probability of selecting the corresponding trial block.
The frequency of trial block activation during wakefulness (\texttt{p\_wake}) is computed based on the selected trial blocks. The probability of activation during sleep (\texttt{p\_sleep}) is computed using a power-law transformation
\begin{equation}
    p_{\text{sleep}} = \frac{p_{\text{wake}}^\gamma}{\sum p_{\text{wake}}^\gamma}
\end{equation}
where \(\gamma = 2.0\) controls the weighting of frequently replayed trials.

\noindent \textbf{Result interpretation:} Figure~\ref{fig:result_1} presents multiple aspects of neural population dynamics and synaptic plasticity, particularly in the context of SpWR event generation and trial block reactivation. The six subplots provide insights into different components of the model. Figure~\ref{fig:result_1}(A) visualizes the evolution of neural activity in a low-dimensional manifold representation. Each trajectory represents a different simulated SpWR event evolving over time. Red crosses indicate the trial templates (predefined), which serve as reference states for trial blocks. The diffusion-based movement in the manifold suggests that neural activity explores the state space stochastically but remains structured. Figure~\ref{fig:result_1}(B) represents neuronal activation patterns across different SpWR events. The x-axis represents SpWR events, while the y-axis represents different neurons. The color scale indicates the activation level (intensity of spiking). The structured variations suggest event-dependent neural activation, reinforcing that neural activity during SpWR is modulated by prior experience. Figure~\ref{fig:result_1}(C) shows the similarity between different SpWR events and predefined trial blocks. Each row corresponds to a different SpWR event, and each column corresponds to a trial block. A high similarity score (yellow) suggests that an SpWR event is highly correlated with a given trial block. The variability in similarity scores suggests differential replay, meaning that certain trial blocks are more strongly reactivated in some events than others. In Figure~\ref{fig:result_1}(D), the blue dots indicate the selected trial blocks (i.e., blocks chosen based on the final manifold state). The red crosses indicate the decoded trial blocks (i.e., blocks inferred based on final trajectory proximity to trial templates). The close alignment between selected and decoded blocks suggests that the model successfully captures the structure of replayed events. Occasional discrepancies indicate some noise in event selection. Figure~\ref{fig:result_1}(E) shows the relative frequency of trial block activations during wakefulness (blue) and sleep (orange). Trial blocks that are frequently active during wakefulness are preferentially reactivated during sleep. This supports the hypothesis that experience-dependent memory consolidation occurs during sleep. Figure~\ref{fig:result_1}(F) shows the distribution of synaptic weights after learning. The y-axis is in log scale, indicating that a large proportion of synaptic weights remain small, while fewer weights grow larger. The presence of a long-tailed distribution suggests selective strengthening of certain connections. This aligns with Hebbian learning, where stronger associations form between co-activated neurons.

\subsection{Interpretation of neural networks}
Our proposed framework is not a neural architecture nor AI models. In stead, we aim to interpret a neural networks by elucidating its neural activity (through SpWR) and inter-neuron connections (via Heider's consistency theory). In other words, we can choose any neural architecture (FNN, RNN, GRN, LSTM, and so on) for this evaluation. In this research, synaptic weight optimization based on the proposed framework, and recurrent neural networks (RNNs) to classify knowledge embeddings mapped onto fixed neuron embeddings. Evaluations on other neural architectures are left for the future work. The pipeline consists of four key stages: (1) \textit{Data preparation} where \( m = 50 \) fixed neuron embeddings are generated and clustered into \( k = 3 \) groups using \( K \)-Means, while \( n_{\text{train}} = 30 \) and \( n_{\text{test}} = 20 \) knowledge embeddings are mapped to the closest neuron groups based on Euclidean distance. (2) \textit{Synaptic weight optimization} where an initial adjacency matrix \( W \) is symmetrized and optimized over \( 200 \) iterations using gradient descent (\( \eta = 0.001 \)) to minimize the balance energy function, enforcing structural balance within neuron triads. (3) \textit{Model definitions} and \textit{training}, where models are implemented to classify knowledge embeddings, with the latter incorporating an additional balance energy penalty in the loss function. The models are trained using Adam optimization (\( \alpha = 0.01 \)) over \( 100 \) epochs, minimizing cross-entropy loss. (4) \textit{Performance evaluation} where training loss and test accuracy are tracked, showing the impact of balance regularization on classification. 

\noindent \textbf{Result interpretation:} Figure~\ref{fig:result_2_2} and ~\ref{fig:result_2_1} shows the interpretation of neural connection using our proposed framework, using RNN, LSTM and GRU (top to bottom). The proposed approach can effectively map the knowledge to model's neurons, showed in Figure~\ref{fig:result_2_2} and ~\ref{fig:result_2_1}(A) of each subplots. We further construct a graph to represent such connections (keeping edge weight in $95^{th}$) percentile (Figure~\ref{fig:result_2_2} and ~\ref{fig:result_2_1}(B)). This interpretation of neurons in DL models is novel to extend of our knowledge. The centroids and distribution of each neural groups are also showed in Figure~\ref{fig:result_2_2} and ~\ref{fig:result_2_1}(E,F). Besides, the balance energy reduces over epochs indicates that the model is learning under the proposed constrains. The distribution of synaptic weights is varying across backbone neural architecture (Figure~\ref{fig:result_2_2} and ~\ref{fig:result_2_1}(D)). However, the decision boundary remains the same (Figure~\ref{fig:result_2_2} and ~\ref{fig:result_2_1}(H)). This means our approach maintains the predictive performance and enhance model interpretability by establishing inter-neural connections.

\section{Discussion} \label{sec:discussion}
\subsection{Rationale of the proposed framework}
Our work's core rationale lies in bridging gaps between biological neural processes and computational models, offering a novel perspective on how structured neural activity and cognitive consistency shape learning and memory. We have used SpWR events as a mechanism for memory replay, aligning with neuroscientific evidence that offline reactivation strengthens synaptic connections. By formalizing SpWR generation and trial block selection (Equations~\ref{eq:spwr_naive}, ~\ref{eq:wakefulness}), our proposed framework explains how experiences are prioritized during wakefulness and sleep, reinforcing important memories while pruning irrelevant ones. This mirrors Hebbian learning principles and synaptic plasticity (Equation~\ref{eq:synaptic_plasticity}), providing a mathematical basis for experience-dependent consolidation. 

The use of low-dimensional manifolds (Equation~\ref{eq:manifold}) to model neural population activity reflects the brain’s efficient encoding of high-dimensional data. This approach captures structured drift in neural representations (Equation~\ref{eq:manifold_dyna}), which is critical for understanding how agents adapt to environments over time. The manifold framework also enables decoding of trial blocks (Equation~\ref{eq:neural_clustering}), linking abstract neural activity to observable behavior.

By incorporating Heider’s consistency theory (Equation~\ref{eq:e_balance}), the model introduces a novel constraint on synaptic interactions, ensuring neural triads maintain structural balance. This bridges social psychology with neuroscience, suggesting that cognitive consistency is not just a high-level phenomenon but a fundamental property of neural circuits. The balance energy function (Equation~\ref{eq:e_balance}) and its gradient (Equation~\ref{eq:e_diff}) enforce stable, coherent memory representations, offering a mechanism for resolving cognitive dissonance at the synaptic level. The framework extends to artificial neural networks (e.g., RNNs, LSTMs) by mapping knowledge embeddings to neurons (Equation~\ref{eq:neural_maps}) and enforcing balance constraints. This enhances interpretability, revealing how synaptic weights and neural clusters encode information while maintaining consistency (Figure~\ref{fig:result_2_2} and ~\ref{fig:result_2_1}). Such insights could inform the design of more robust, neuromorphic computing systems.

\subsection{Limitations of the proposed work}\label{sec:limitation}
While innovative, our research faces several limitations that warrant further exploration. First, the model abstracts neural activity into firing rates and low-dimensional manifolds, potentially oversimplifying the brain’s heterogeneous, nonlinear dynamics. For instance, place cell drift (Equation~\ref{eq:neural_activity}) assumes Gaussian noise, but real neural variability may involve non-Gaussian or non-stationary processes. Second, the balance energy function (Equation~\ref{eq:e_balance}) focuses on triadic interactions, neglecting higher-order network effects. Third, the effect of parameters on performance is not fully addressed in this research. The model relies on hyperparameters like \(\beta\) (replay sharpness), \(\gamma\) (balance weight), and \(\eta\) (learning rate). Their optimal values are not derived from first principles but chosen empirically, raising questions about robustness. For example, excessive \(\gamma\) might over-penalize imbalance, stifling adaptive plasticity. Finally, the balance theory assumes symmetric synaptic weights (Equation~\ref{eq:synatic_weight}), which simplifies the directed, asymmetric connections common in biological networks. This may limit the model’s applicability to systems with hierarchical or modular architectures.

\section{Conclusion} \label{sec:conclusion}
To this end, we outline several promising directions for future research to extend this work. A straightforward improvement involves addressing the model’s limitations discussed in Section~\ref{sec:limitation}. Additionally, this framework could be applied to real-world AI-driven applications, such as image classification \cite{nguyen2021csnas,zanddizari2021new}. Another notable direction is leveraging this framework to investigate the topological structure of quantum circuits in quantum machine learning \cite{nguyen2022bayesian,nguyen2022quantum,nguyen2024biomarker,nguyen2024duality,nguyen2024quantumETA}. Furthermore, our experimental observations suggest that the dynamics of SpWR exhibit similarities to quantum wavefunction behavior during tunneling through an energy barrier. Therefore, incorporating my previous model on quantum tunneling \cite{nguyen2024quantumtunneling} as an alternative to Section~\ref{subsec:math_spwr} is a worthwhile avenue for exploration. Finally, biologically inspired neural architectures featuring categorized neurons—such as sensory neurons, interneurons, and motor neurons—present a compelling candidate for future evaluations.

\bibliography{ref}
\bibliographystyle{abbrv}

\end{document}